\documentclass[a4paper,10pt]{article}
\usepackage[margin=1in]{geometry} 
\usepackage[utf8]{inputenc} % allow utf-8 input
\usepackage[T1]{fontenc}
\usepackage{graphicx} 
\date{\today}
\usepackage{amsmath, amssymb} 
\usepackage{float} 
\usepackage{hyperref}
\usepackage{authblk} % For author and affiliation formatting
\usepackage{setspace}
\usepackage{caption}
\usepackage{cite}
\usepackage{url}            % simple URL typesetting
\usepackage{booktabs}       % professional-quality tables
\usepackage{amsfonts}    % blackboard math symbols
\usepackage{nicefrac}       % compact symbols for 1/2, etc.
\usepackage{microtype}      % microtypography
\usepackage{xcolor}         % colors
\usepackage{wrapfig}     
\usepackage{graphicx}
\usepackage{subfigure}
\usepackage{bm}

\usepackage{caption}
\usepackage[linesnumbered,ruled,vlined]{algorithm2e}
\title{GraphTorque$:$ Torque-Driven Rewiring \\ Graph Neural Network}

\author{
	Sujia Huang$^{1}$,  Lele Fu$^{2}$,  Zhen Cui$^{3}$, Tong Zhang$^{1*}$, Na Song$^{4}$, Bo Huang$^{1}$ \\
	$^1$Nanjing University of Science and Technology, Nanjing, China \\
	$^2$Sun Yat-Sen University, Guangzhou, China\\
	$^3$Beijing Normal University, Beijing, China \\
	$^4$Putian University, FuZhou, China \\
	\texttt{hsujia2021@163.com}, \texttt{fulle@mail2.sysu.edu.cn}\quad \\
	\texttt{zhen.cui@bnu.edu.cn},
	\texttt{tong.zhang@njust.edu.cn},  \\
	\texttt{ptusn@ptu.edu.cn}, \texttt{huangbo@njust.edu.cn}
}

\date{}

\begin{document}

% Make the title
\maketitle

\begin{abstract}
	Graph Neural Networks (GNNs) have emerged as powerful tools for learning from graph-structured data, leveraging message passing to diffuse information and update node representations.
	However, most efforts have suggested that native interactions encoded in the graph may not be friendly for this process, motivating the development of graph rewiring methods.
	In this work, we propose a torque-driven hierarchical rewiring strategy, inspired by the notion of torque in classical mechanics, dynamically modulating message passing to improve representation learning in heterophilous and homophilous graphs.
	% Specifically, we define an interference-aware torque metric that integrates pairwise distance and energy attributes, guiding each node to tend to aggregate information from its nearest low-energy neighbors.
	Specifically, we define the torque by
	treating the feature distance as a ``lever arm vector” and the neighbor feature as a ``force vector”
	weighted by the homophily disparity between nodes.
	We use the metric to hierarchically reconfigure each layer’s receptive field by judiciously pruning high-torque edges and adding low-torque links, suppressing the impact of irrelevant information and boosting pertinent signals during message passing.
	Extensive evaluations on benchmark datasets show that the proposed approach surpasses state-of-the-art rewiring methods on both heterophilous and homophilous graphs.
\end{abstract}

\section{Introduction}
Graph-structured data composed of vertices and edges encode entities and their relationships.
Graph neural networks (GNNs) have emerged as a powerful framework for processing such data, with widespread applications in biomolecular modelling \cite{gligorijevic2021structure,xia2023understanding}, recommendation systems \cite{chen2024macro,anand2025survey} and beyond \cite{jiang2023pdformer,liu2025vul,huang2025boosting}.
% By iteratively propagating and aggregating node features across the graph, GNNs capture both local connectivity and global topology, obtaining informative representations from complex graphs.
% Typical tasks in these domains include graph classification, node classification and link prediction \cite{}, where we focus on 
% As one of the typical tasks in these domains, node classification \cite{} aims to accurately classify samples into clusters they belong to
At the heart of GNNs lies message passing, which iteratively propagates and aggregates information along edges to enrich node representations.
Therefore, the graph structure not only encodes entity interactions but also critically determines model performance \cite{zhang2020dynamic,YangYBZZGXY23,DBLP:conf/iclr/QianMAZBN024}. 

In practice, however, raw graphs frequently harbour spurious or missing links arising from noise or sampling artefacts, compromising their effectiveness as substrates for message propagation.
In response, recent work has devised diverse graph rewiring strategies that selectively remove and add edges to optimize message passing and boost predictive accuracy \cite{battaglia1806relational,xue2023lazygnn,abboud2022shortest,DHGR,bose2025can}.
Such dynamic topology adjustment is crucial not only for mitigating spurious connections but also for addressing heterophily, where nodes with dissimilar labels or features tend to be connected \cite{yang2021diverse,zheng2023finding,lee2023towards}.
In such scenarios, homophily‑based GNNs can be misled by abundant heterophilous connections, yielding entangled representations and degraded classification accuracy.

%	Graph rewiring algorithms fall into two broad classes.
%	Preprocessing approaches operate on the graph prior to model training:
%	predefined strategies are applied to prune or augment edges, producing a static topology that alongside the original node features is subsequently supplied to GNNs \cite{topping2021understanding,nguyen2023revisiting,xue2023lazygnn}.
%	%For instance, \cite{JDR} used node features to boost graph, jointly performing the graph rewiring and the feature denoising.
%	While these methods deliver clear interpretability and have demonstrated strong results, their dependence on a static topology limits each node’s capacity to adaptively integrate information. 
%	By contrast, end‑to‑end approaches integrate graph rewiring directly into the training loop, allowing gradient signals from the downstream task to dynamically reshape graph structures \cite{rong2019dropedge,giraldo2023trade,qian2023probabilistically}.topping2021understanding,giraldo2023trade

\begin{wrapfigure}{r}{0.5\textwidth}
	\centering
	\begin{minipage}{0.5\textwidth}
		\centering
		\includegraphics[width=\linewidth]{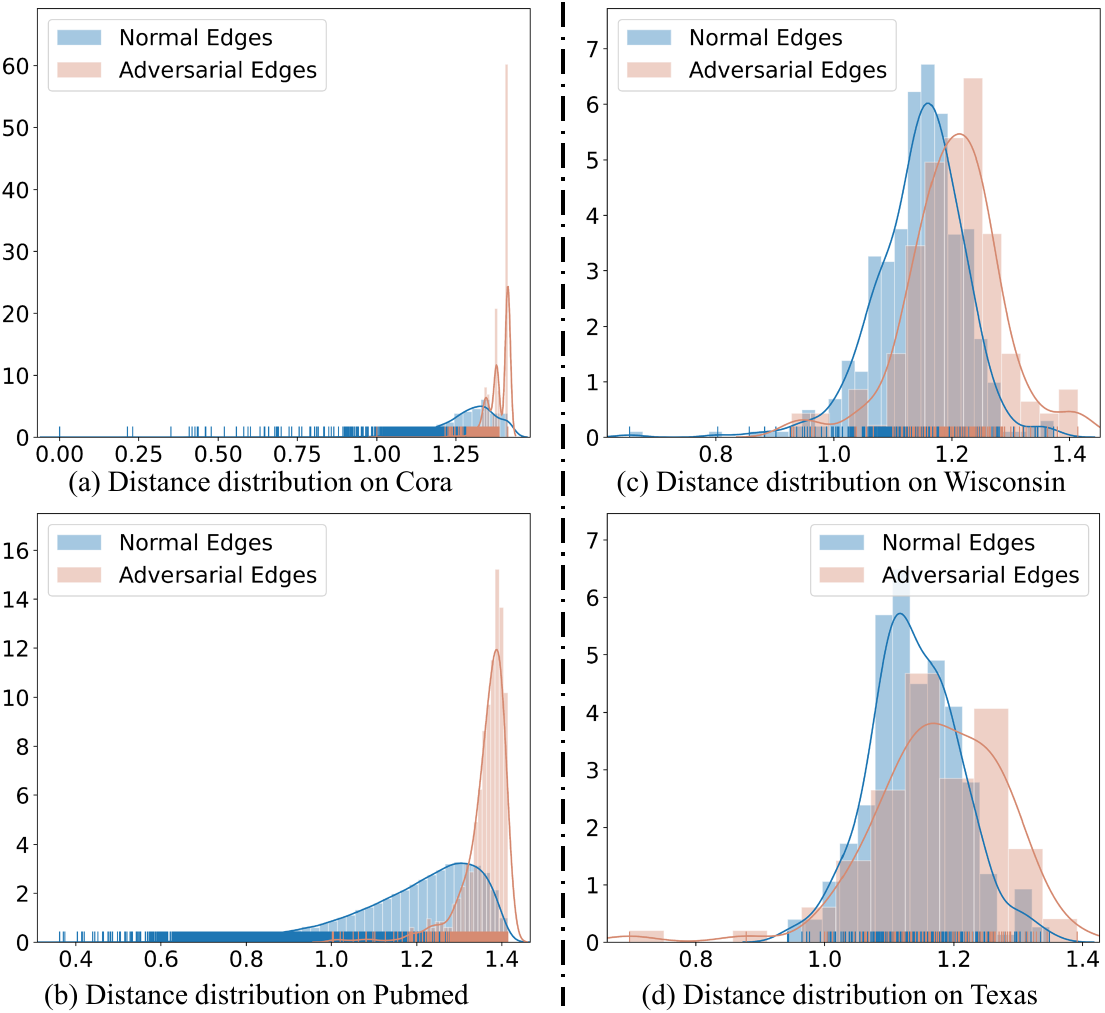}
	\end{minipage}
	\caption{Density distributions of distances for normal vs. adversarial edges on homophilous graphs (a) Cora and (b) Pubmed and heterophilous graphs (c) Wisconsin and (d) Texas.}
	\label{fig:kde}
\end{wrapfigure}

One of the core challenges in graph rewiring is quantifying the impact of edges on message passing. 
A key factor in this process is the similarity between node pairs, often measured using the Euclidean distance, a commonly used metric for assessing similarity. 
In general, the larger the distance between nodes, the weaker their interaction strength, and the less useful information can be transmitted, as supported by previous studies that employed node similarity as a proxy for edge weights \cite{wang2020gcn, zhou2024blockgcn}. 
To intuitively observe this, we simulate adversarial attacks by injecting adversarial edges into raw graphs and visualize the distance distribution of the edges, enabling us to examine whether adversarial and original edges exhibit distinct distributional patterns.
As shown in Fig. \ref{fig:kde}(a)–(d), the distribution trends in both homophilous datasets (Cora and PubMed) and heterophilous datasets ( Wisconsin and Texas ) consistently indicate that adversarial edges (in red) tend to connect node pairs with larger feature distances.
These observations suggest that adversarial attacks preferentially create long-range links so that they disrupt message passing at their target nodes.
Furthermore, we observe that normal edges in heterophilous datasets also exhibit a distribution skewed toward larger distances, more pronounced than in homophilous datasets. This is because heterophilous graphs contain a much higher proportion of heterophilous edges, which typically span node pairs with low similarity (i.e., large distances).
Given that a minority of long-range neighbors can convey crucial information while nearby neighbors may propagate misleading signals, the feature quality of neighboring nodes should be another key factor in assessing edge significance.

This brings to mind the concept of \textbf{\textit{Torque}} in classical mechanics, which is mathematically defined as the cross product of a lever arm (the position vector from the axis of rotation to the point of force application) and a force.
Recently, torque has found applications in fields such as biology \cite{tang2023bipolar, dzhimak2022genesis, drobotenko2025medium} and spintronics \cite{kovarik2024spin, camarasa2024spin}.
Heuristically, we extend this concept to graphs by treating the distance vector between nodes as the lever arm and the feature vector of a neighboring node as the force. 
Their product yields a graph torque, which measures an edge’s negative impact: higher torque flags greater interference.
To our knowledge, this is the first work to integrate a physics-inspired torque into graph rewiring, enabling an interference-aware message passing.

% In this paper, we devise a \underline{T}orque‑driven Hierarchical \underline{R}ewiring strategy (THR) for GNNs (\underline{TR}GNN), based on the intuitive principle that low‑energy nodes should preferentially connect to other low‑energy nearest neighbors.
Specifically, we devise a \underline{T}orque‑driven \underline{H}ierarchical \underline{R}ewiring strategy (THR) for GNNs, which dynamically refines message passing to excel in both homophilous and heterophilous graph.
In THR, each edge is assigned a torque value that quantifies its interference strength, with larger torques indicating less reliable connections.
Sepcifically, we define torque by treating the difference between node representations as a ``lever arm vector”, which emphasizes long-range or heterophilous links.
Meanwhile, the neighbor feature is treated as a ``force vector” weighted by the disparity in the homophily ratios between node pairs. 
This disparity captures the difference in their local label homophily, which has been theoretically shown to jointly influence the expressive power of GNNs together with feature distance. %that together with the feature distance to the model performance
%Their product defines a torque to quantify the interference strength of edges, where larger torques flag less reliable connections.
Leveraging this torque value, THR hierarchically reconfigures each layer’s receptive field via automatically removing undesirable edges that degrade performance and introducing low-torque significant connections, thus effecting interference-resistant and importance-aware propagation.
This rewiring is performed end-to-end, where message passing operates on the continuously updated graph, while the evolving node representations enhance torque computation.

\textbf{Contributions:} 1) To the best of our knowledge, we are the first to apply the concept of torque from physics to graph rewiring, resulting in THR, which enhances GNNs' resilience to both homophily and heterophily.
2) We propose a hierarchical rewiring strategy that adaptively determines each layer's receptive field by automatically pruning undesirable connections and adding significant edges.
3) Comprehensive  experiments indicate that THR improves the performance of various GNNs and outperforms existing state-of-the-art rewiring strategies.

%	Our contributions are as follows,
%	\begin{itemize}
	%		\item We propose a torque-driven rewiring graph neural network, which adaptively infers the receptive field suited for downstream tasks via an end-to-end manner, substantially enhancing GNNs’ resilience to both heterophily and noise.
	%		\item We develop a importance-aware metric that fuses Euclidean distance with learned node‑energy scores to promote the hierarchical rewiring through automatically pruning undesirable connections and adding significant edges. %across multiple neighborhood orders.
	%		\item We conduct extensive experiments to indicate that TorqueGNN outperforms state-of-the-art competitors on both heterophilous and homophilous benchmarks and works robustly on graphs with noise. 
	%		The hierarchical rewiring mechanism surpasses other rewiring strategies.
	%	\end{itemize}

\section{Related Work}

% The inherently local scope of standard message passing fails to capture long‑range dependencies.
Standard message passing in GNNs, which aggregates information from local neighbourhoods, struggles to capture long-range dependencies.
A common remedy is to stack multiple layers to expand the receptive field \cite{SGC,GCNII,xu2025few}, but this approach frequently encounters fundamental limitations such as over‑smoothing and over‑squashing.
To address these bottlenecks, graph rewiring techniques have recently emerged as a powerful strategy to restructure connectivity and enhance information flow. 
% Graph rewiring has recently emerged as a powerful solution. 
For example, Expander GNNs and ExPhormer perform graph rewiring by merging multi‑hop neighbourhoods or injecting virtual nodes \cite{deac2022expander,gabrielsson2023rewiring,shirzad2023exphormer}.
\cite{fosr} adds edges based on spectral expansion to mitigate over-smoothing and over-squashing, while
degree-preserving local edge-flip algorithms are developed \cite{banerjee2022oversquashing}.
\cite{topping2021understanding,di2023over} analyze the root causes of over-squashing, demonstrating that both spatial and spectral rewiring can effectively counteract this bottleneck. %by reweighting or reconfiguring edges. 
%Despite these advances, real-world graphs often remain inherently noisy, limiting the reliability of message passing.

Moreover, \cite{FAGCN} highlights the challenge posed by heterophilous edges, wherein the aggregation of dissimilar node signals can lead to entangled representations and misclassifications. 
% Raw graph structures are frequently suboptimal substrates for message passing, especially for heterophily graphs, numerous heterophilous edges can induce confusing information, thereby limiting the expressive power of GNNs \cite{topping2021understanding,giraldo2023trade}. 
To mitigate the effect of such undesirable connections, %\cite{JDR} simultaneously perform graph rewiring and feature denoising to boost classification accuracy.
\cite{DHGR} compares the neighborhood feature distribution and neighborhood label distribution between node pairs to pruning heterophilous edges and adding homophilous edges.
%There are many methods specifically devised for heterophily.
%For example, 
\cite{GGCN,ACMGNN,LRGNN} employ signed message propagation, assigning positive weights to homophilous links and negative weights to heterophilous ones.
This enables differentiated updates on the heterophilous graphs, thereby amplifying similarity among homophilous nodes while suppressing similarity among heterophilous ones. 
However, \cite{M2MGNN} has shown that, although single‑hop signed adjacency matrix aids in distinguishing features of different classes, the multi‐hop propagation matrix introduced to expand the receptive field often degrades performance.

We are inspired by the torque in physics to design a new rewiring mechanism that hierarchically eliminates undesirable connections and incorporates task‑relevant edges. 
By dynamically reshaping the receptive field during training, our method enhances the discriminative power of GNNs on both homophilous and heterophilous graphs.
% \subsection{Graph Rewiring}

% \subsection{Energy-based Model}

\section{Preliminaries}
\subsection{Notations}
Let us define an undirected graph dataset as $\mathcal{G} = (V, \mathcal{E})$, comprising $N$ nodes $\{v_i\in V\}_{i=1}^N$ and $K$ edges $\{e_{k}\doteq \langle i,j \rangle\in{\mathcal{E}}\}_{k=1}^K$, where each edge $k$ encodes a connection between nodes $v_i$ and $v_j$.
We denote the adjacency matrix by $\mathbf{A}\in \{0, 1\}^{N\times N}$, where $A_{\langle i,j \rangle} = 1$ iff nodes $v_i$ and $v_j$ are connected, 0 otherwise.
Furthermore, $\widehat{\mathbf{A}}=\mathbf{A}+\mathbf{I}$ indicates $\mathbf{A}$ with added self-loops, and $\widetilde{\mathbf{A}} = \widehat{\mathbf{D}}^{-1/2}\widehat{\mathbf{A}}\widehat{\mathbf{D}}^{-1/2}$ denotes the symmetrically normalized adjacency matrix with $\widehat{{D}}_{\langle i,i \rangle}=\sum_{i=1}^N\widehat{A}_{\langle i,j \rangle}$. 
Each node is associated with a feature vector, and we write $\mathbf{X}\in\mathbb{R}^{N\times d}$ for the node feature matrix, where the $i$-th row $\mathbf{x}_i \in \mathbb{R}^d$ represents the $d$-dimensional features of node $v_i$. 
Among $N$ nodes, $N_{lab}$ nodes are labeled, with ground-truth labels encoded in a matrix $\mathbf{Y}\in\mathbb{R}^{N_{lab}\times c}$, where each row $\mathbf{y}_i$ is a one-hot vector indicating the class label among $c$ categories.

\subsection{Message Passing}
Consider a graph with adjacency matrix $\mathbf{A}$ and node feature matrix $\mathbf{X}$.
Message passing in a GNN proceeds by iteratively propagating and aggregating neighborhood information as
\begin{equation}\label{eq:mp}
	\mathbf{h}_i^{(l+1)} = \text{Upd}\big(\mathbf{h}_i^{(l)}, \sum_{v_j\in\mathcal{N}_{i}}\text{Agg}(\mathbf{h}_j^{(l)}, A_{\langle i,j \rangle})\big),
\end{equation}
where $\mathbf{h}_i^{(0)} = \mathbf{x}_i$, and $\mathbf{h}_i^{(l+1)} \in \mathbb{R}^m$ is the representation of node $v_i$ in the $(l+1)$-th layer. 
$\text{Agg}(\cdot)$ computes the incoming message from a neighbor $v_j$, and $\text{Upd}(\cdot)$ updates the representation of node $v_i$.
Rather than relying on the raw adjacency matrix $\mathbf{A}$, most GNNs adopt a modified propagation operator $\mathcal{A}$.
For example, GAT \cite{GAT} replaces each non-zero entry of $\mathbf{A}$ with a learned attention coefficient that depends on the representations of the corresponding node pair.

\subsection{Node-level Homophily and Heterophily}
For a set of nodes with labels, the homophily ratio of each node quantifies the tendency of the node to share the same label as its neighbors.  
Considering a node $v_i$ and its set of neighbors $\mathcal{N}_i$, the homophily ratio $h_i^+$ of $v_i$ is defined as: $h_i^+ = \frac{|\{\mathbf{y}_i=\mathbf{y}_j|v_j\in \mathcal{N}_i\}|}{|\mathcal{N}_i|}$. 
The value of $h_i^+$ lies in the range $[0,1]$, where values closer to 1 indicate a higher degree of homophily (or lower heterophily), while values nearer to 0 signify the opposite.
To quantify the homophily of the entire graph $\mathcal{G}$, we compute the average homophily across all nodes: $\mathcal{H}(\mathcal{G}) = \frac{\sum_{i=1}^{N}h_i^+}{N}$. 
%Many previous works have explored heterophily using the above node-level homophily metric and have proposed various approaches, such as signed edges, to reduce the impact of confusing information brought by non-similar neighbors \cite{GGCN,GPR-GNNs}.

\section{Methodology}
% Real‑world graphs are often marred by spurious or missing links arising from noise, leading to a unfriendly computational graph for message passing.
% This challenge is compounded in highly heterophilous networks, as there are more heterophilous edges so that inter-class representations are indistinguishable.
% To counteract this, \cite{GGCN,ACMGNN,FAGCN} employ signed message propagation, assigning positive weights to homophilous links and negative weights to heterophilous ones.
% This enables differentiated updates on the heterophilous graphs, thereby amplifying similarity among homophilous nodes while suppressing similarity among heterophilous ones. 
% However, \citet{M2MGNN} have shown that, although single‑hop signed adjacency matrices $\mathbf{A}^{(l)}$ aids in distinguishing features of different classes, the multi‐hop propagation matrix $\prod_{l=1}^L\mathbf{A}^{(l)}$ introduced to expand the receptive field often undermines performance.
In this section, we propose a novel graph rewiring strategy that unfolds in three key stages: (i) computing edge torques, (ii) rewiring propagation matrix, and (iii) adjusting message passing.
The full algorithmic pseudocode is provided in Appendix~\ref{sec:alg}.
%while a detailed complexity analysis of TorqueGNN is included in Appendix~\ref{sec:com_ana}.

\subsection{Derive Graph Torque}\label{sec:two_pro}
In classical mechanics, \textit{\textbf{torque}} is defined as the vector cross product of a force and its lever arm:
\begin{equation}
	\mathbf{T} = \mathbf{r} \times \mathbf{F}, \quad |\mathbf{T}| = |\mathbf{r}||\mathbf{F}|\sin\theta,
\end{equation}
where $\mathbf{r}$ denotes the displacement vector, $\mathbf{F}$ indicates the force vector, and $\theta$ is the angle between them. 
The magnitude of the torque directly governs an object’s tendency to rotate under the applied force.
In GNNs, the systematic exploration of node interactions can mirror the lever arm–force relationship underlying torque in classical mechanics.
Specifically, torque on a graph may be conceptualized by treating the displacement vector $\mathbf{D}_{\langle i, j \rangle}$ between a central node $v_i$ and its neighbor $v_j$ as the effective “lever arm”, while the neighbor’s features $\mathbf{x}_j$ act as the applied ``force".
However, the contribution of this force varies across different central nodes.
Recent studies \cite{DBLP:conf/nips/MaoC0H0ZST23, huang2025boosting} demonstrate that the generalization of GNNs is influenced by two key factors: the proximity of aggregated features and the disparity in homophily ratio, with smaller values yielding better generalization.
Inspired by this, we introduce the homophily ratio disparity term $E_{\langle i, j \rangle}$ to modulate the effective force, thereby not only capturing the heterogeneous influences of neighboring nodes but also unifying these two factors within the torque formulation to reduce generalization error.
%This joint consideration of lever arm and disparity-adjusted force allows the torque formulation to capture these two critical factors simultaneously, yielding more reliable graph optimization.

Mathematically, for an edge $k$ connecting nodes $v_i$ and $v_j$, we define the corresponding torque as follow,
\begin{equation}\label{eq:graphTorque}
	\mathbf{T}_{e_k} = \mathbf{D}_{\langle i, j \rangle} \times E_{\langle i, j \rangle}\mathbf{x}_j.
\end{equation}
Its magnitude, denoted $T_{e_k}$, measures the disturbance imposed by the message passing along edge $e_k$ on node $v_i$.
The value naturally increases with larger distances or higher structural disparity, with edges maximizing both factors yielding the greatest torque value that represents the highest priority for graph rewiring.
A central goal is therefore to provide a principled definition of the displacement vector  $\mathbf{D}_{\langle i, j \rangle}$ and the homophily ratio disparity $E_{\langle i, j \rangle}$ in Eq. \ref{eq:graphTorque}, after which we detail their construction.
% \textbf{Define Two Metrics of Connections.}

\textbf{Metric 1: Displacement Vector.}
To mitigate the effect of noise in raw graphs and features, we estimate the displacement vector $\mathbf{D}_{\langle i,j \rangle}$ using optimized node representations and compute the pairwise distances as
\begin{equation}\label{eq:distance}
	\mathbf{D}_{\langle i,j \rangle} = \mathbf{h}_i - \mathbf{h}_j, D_{\langle i,j \rangle} = \|\mathbf{h}_i - \mathbf{h}_j\|_2,
\end{equation}
where $\mathbf{h}_i=\text{gCov}(\mathbf{x}_i,\mathbf{A}; \mathbf{\Theta})$\footnote{$``\text{gCov}"$ can be instantiated with any standard GNN layer, such as GCN, GPRGNN, or APPNP.} denotes the representation of $v_i$  obtained via a graph convolution operator ``gCov" parameterized by $\mathbf{\Theta}$ and followed by a ReLU activation.
% $\mathbf{h}_i^{(0)}=\mathbf{x}_i$  Furthermore, $\mathcal{A}$ is the refined propagation matrix for the $l$-th layer.

\textbf{Metric 2: Homophily Ratio Disparity.} 
%While previous work highlights the importance of feature distance between node pairs, 
Considering that recent studies emphasize the importance of capturing the homophily ratio disparity in addressing heterophilous graphs, we weight the neighbor features using this disparity to incorporating  both it and distance into the torque formulation.
To estimate node-level homophily, it is crucial to annotate the labels of neighboring nodes around a given node.
Since labeled data are often scarce, we utilize the model's final outputs to generate pseudo-labels for unlabeled nodes, with their accuracy improving as the model is progressively optimized.
Formally, $E_{\langle i, j \rangle}$ is computed by
\begin{equation}\label{eq:edge_energy}
	E_{\langle i, j \rangle} = |h_i^+ - h_j^+|, \ 
	h_i^+ = \frac{|\{v_j| \hat{\mathbf{y}}_i = \hat{\mathbf{y}}_j, v_j \in \mathcal{N}_i\}|}{|\mathcal{N}_i|}.
\end{equation}
Here, $\hat{\mathbf{y}}_i$ denotes the ground-truth label for labeled nodes or the pseudo-label for unlabeled nodes.
Finally, the torque value of edge $e_k$ is computed by:
\begin{equation}\label{eq:graphTorque_value}
	T_{e_k} = \|\mathbf{D}_{\langle i, j \rangle} \times (E_{\langle i, j \rangle}\mathbf{h}_j)\|_2 = \sqrt{D_{\langle i, j \rangle}^2\cdot (E_{\langle i, j \rangle}\|\mathbf{h}_j\|_2)^2 - (E_{\langle i, j \rangle}\mathbf{D}_{\langle i, j \rangle}\cdot   \mathbf{h}_j)^2} \footnote{This form follows directly from the vector identity $\|\mathbf{a}\times \mathbf{b}\|^2 = \|\mathbf{a}\|^2 \|\mathbf{b}\|^2 - (\mathbf{a}\cdot \mathbf{b})^2$.} %The angular dependence (sin $\theta$) is thus implicitly captured by the dot product, avoiding to introduce $\theta$ explicitly.},
\end{equation}
This formulation captures the combined effects of distance and disparity, facilitating a physics-inspired approach to graph rewiring.

\subsection{Adjust Message Passing}
%\subsubsection{Hierarchical Graph Rewiring}

\textbf{Edge-removal High-order Rewiring.}
Herein, we propose an automated threshold learning mechanism that identifies the optimal number of edges to prune by pinpointing the largest successive torque gap. 
Specifically, we first rank all $K$ edges in descending order of their torque values to form a torque‐sorted list (TSL), denoting its $k$-th entry as $\widetilde{e}_k$ with torque $\widetilde{T}_{e_k}$, so that $\widetilde{T}_{e_1}\geq\widetilde{T}_{e_2}\geq\cdots\geq\widetilde{T}_{e_K}$.
We then calculate the torque gap between two consecutive links by
\begin{equation}\label{eq:gap}
G_{k,k+1} = \mu_k\times \frac{\widetilde{T}_{e_k}}{\widetilde{T}_{e_{k+1}}+\delta}, %\quad \widetilde{T}_{k+1}\neq 0.
\end{equation}
where $\delta$ is a small constant to prevent division by zero, and the weight $\mu_k$ reflects the proportion of anomalous edges, those whose distance $D$, disparity $E$ and torque $T$ all exceed their respective means, that are captured within the top $k$ torque-ranked set, emphasizing the boundary between desirable and undesirable connections.
The computation formula of $\mu_k$ is defined as 
\begin{equation}\label{eq:weight}
\begin{split}
	\mu_k &= \frac{|High\_e\cap Top\_k|}{|High\_e|},\\
	\hfill
	High\_e &= \{e_k\doteq {\langle i,j \rangle}|D_{\langle i,j \rangle}\geq \Bar{D}, E_{\langle i,j \rangle}\geq \Bar{E}, T_{\langle i,j \rangle}\geq \Bar{T}, \langle i,j \rangle\in\mathcal{E}\},\\
	Top\_k &= \{e_k\doteq {\langle i,j \rangle}|Top_k\{T_{\langle i,j \rangle}\},\langle i,j \rangle\in\mathcal{E}\},
\end{split}
\end{equation}
where $\Bar{D}, \Bar{E}, \Bar{T}$ denote the mean values of distance, disparity and torque, respectively, computed over all $K$ edges.
The set $High\_e$ comprises edges exhibiting above-average values across all three metrics, while $Top\_k$  contains the top $k$ connections in TSL.
According to Eq. \ref{eq:gap}, we can identify the optimal cutoff by locating the largest torque gap $\mathcal{K} = \arg\max\limits_{0\leq k\leq K-1}G_{k, k+1}$, which separates the edge set into two groups: undesirable connections ($\widetilde{e}_1,\cdots,\widetilde{e}_{\mathcal{K}}$) and desirable connections ($\widetilde{e}_{\mathcal{K}+1},\cdots,\widetilde{e}_{K}$).

In practice, multi-layer GNNs, such as APPNP \cite{APPNP} and GCNII \cite{GCNII}, are widely adopted to enlarge the receptive field of graph convolutions.
To allow each layer to adapt the adjacency relationships based on the current information and focus on different structural features of the graph, we adopt a hierarchical rewiring strategy.
Building on the pairwise torque gap formulation introduced above, we extend this mechanism across multiple propagation layers.
In specific, for each layer $l$, we construct a dedicated propagation matrix that enables selective filtering of undesirable high-order interactions.
Let $\mathcal{A}^{(l)}$ denote the refined propagation matrix used in $l$-th layer and $\mathcal{A}^{(0)}=\mathbf{A}$.
We identify the current neighbors $v_j$ of a given node $v_i$ based on $\mathcal{A}^{(l)}$, and recompute the torque as follows
\begin{equation}\label{eq:l_rwo_properties}
\mathbf{T}_{\langle i,j \rangle}^{(l+1)} = (\mathbf{h}_i^{(l)} - \mathbf{h}_j^{(l)}) \times E_{\langle i,j \rangle}\mathbf{h}_j^{(l)},
\end{equation}
where $l=0, \cdots, L-1$.
%The initial representation $\mathbf{h}_i^{(0)}=\mathbf{x}_i\mathbf{\Theta}$ is computed via a linear transformation of the input feature $\mathbf{x}_i$.
Consequently, we gain the $(l+1)$‑th order torque $T^{(l+1)}$ and the corresponding gap $G^{(l+1)}$ using Eqs. \ref{eq:graphTorque_value}-\ref{eq:weight}, from which we derive a pruned propagation matrix $\mathcal{A}^{(l+1)^*}$ with $(K-\mathcal{K})$ non-zero elements. 

\textbf{Edge-addition High-order Rewiring.}
In the previous steps, we remove undesirable neighbors by computing the torque of existing edges based on two key attributes.
Extending this strategy, we also consider expanding the receptive field by adding edges that are initially absent but potentially beneficial for message passing. 
However, evaluating torque across all missing edges is computationally intractable, so that we construct a candidate set $\mathcal{T}$ by selecting, for each node, its top-$t$ most similar peers.
We then compute the $(l+1)$-th order torque $T^{(l+1)}$ for the resulting $N\times t$ candidate edges, and select $r\times N \times t$ edges with the lowest torque values, where $r$ is a sampling ratio.
Nevertheless, this hard selection process is inherently non-differentiable and thus cannot be used in gradient-based optimization. 
To overcome this, we adopt the Gumbel-Softmax reparameterization trick \cite{DBLP:conf/iclr/JangGP17}, which enables differentiable sampling by approximating discrete decisions with a continuous relaxation.
For each candidate edge $k$, we define its logits $\bm{\pi}_k=[\pi_{k0}, \pi_{k1}]$, where $\pi_{k0}=T_{e_k}^{(l+1)}$ (discard) and $\pi_{k1}=1-T_{e_k}^{(l+1)}$ (select).
Drawing independent noise $g_{kj}\sim \text{Gumbel (0, 1)}$, the soft selection probabilities are computed via
\begin{equation}\label{eq:gumbel_softmax}
p_{kj}=\frac{\exp \left(\frac{\log \left(\pi_{kj}\right)+g_{kj}}{\tau}\right)}{\sum_{m=1}^2 \exp \left(\frac{\log \left(\pi_{km}\right)+g_{km}}{\tau}\right)}, \forall j=0,1, k \in\{1,2, \ldots, N\times t\},
\end{equation}
where $\tau$ is a temperature parameter controlling the sharpness of the Gumbel-Softmax distribution.
$p_{k1}$ serves as a differentiable weight indicating the likelihood of selecting candidate edge $k$.
Finally, we construct the rewired propagation matrix $\mathcal{A}^{(l+1)}$ by augmenting $\mathcal{A}^{(l+1)^*}$ with these probabilistically weighted candidate edges, followed by the standard renormalization procedure.
% For high-order representations, the output at the $(l+1)$-th layer is then updated as:
% $\mathbf{h}_i^{(l+1)}=\text{ReLU}\big(\sum\limits_{j\in\{v_i\}\cup\{v_j|\mathcal{A}_{\langle i,j \rangle}^{(l+1)}\neq 0\}}\alpha\mathbf{h}_j^{(l)}+(1-\alpha)\mathbf{h}_i^{(0)}\big)$.

%Then, we can select edges by using $\pi_{k2}$ as the weight of edge $k$, which evolves $\mathbf{A}^{l^*}$ into the rewired propagation matrix $\mathcal{A}^{(l)}$ by adding the new edges and then applying renormalization trick.

\textbf{Messaging Passing on Rewired Graph.}
% To avoid misleading representations in the early stages of training, we refrain from directly sharing the rewired graph across layers, as this would either discard the corresponding neighbor’s information or propagate erroneous signals. 
%      Instead, rewiring at each layer is always performed with respect to the original input graph.
To avoid misleading representations in the early stages of training, which could either discard important neighbor information or propagate erroneous signals, rewiring at each layer is always performed with respect to the original input graph.
By rewiring the adjacency matrix $\mathbf{A}$ as 
described, each propagation layer is endowed with an expanded receptive field capable, enabling the capture of effective multi-level interactions.
%At the $l$-th layer, the model aggregates heterogeneous messages of varying effective orders: neighbors whose connections persist up to the $(l+1)$-th layer contribute $(l+1)$-order information, whereas those whose edges are pruned earlier provide lower-order signals, specifically of order $(l-m)$ if the edge is last retained at the $m$-th layer.

To evaluate the effectiveness of the proposed THR in capturing high-order information in multi-layer GNNs, we use the deep-based, APPNP, as an example. Subsequent ablation studies and parameter analyses are  conducted within this framework.
Let $\mathcal{N}_i^{(l+1)} = \{v_j|\mathcal{A}_{\langle i,j \rangle}^{(l+1)}\neq 0\}$ denotes the refined $(l+1)$-layer neighborhood of node $v_i$; then the forward propagation at the $(l+1)$-th layer of APPNP can be reformulated as:
\begin{equation}
\begin{aligned}\label{eq:intra_rep_updat}
	\mathbf{h}_i^{(l+1)} 
	&=
	\text{ReLU}\Big(\sum\limits_{j\in\{v_i\}\cup\mathcal{N}_i^{(l+1)}}\alpha\mathbf{h}_j^{(l)} +(1-\alpha)\mathbf{h}_i^{(0)}\Big).
	% \text{ReLU}\Big(\sum\limits_{j\in\{v_i\}\cup\mathcal{N}_i^{(l+1)}}\alpha\big(\mathbf{h}_j^{(l)} + \sum_{m=0}^{l-1}\sum\limits_{j \in \mathcal{N}_i^{(m)} \backslash \mathcal{N}_i^{(m+1)}}\mathbf{h}_j^{(l-m)}\big)+(1-\alpha)\mathbf{h}_i^{(0)}\Big).
\end{aligned}
\end{equation}
Here, $\alpha$ controls the trade-off between the hidden representation and the residual connection.
The initial representation $\mathbf{h}_i^{(0)}=\mathbf{x}_i\mathbf{\Theta}$ is computed through a linear transformation of the input feature $\mathbf{x}_i$.
The final node representations from the last layer are passed through a fully connected layer parameterized by $\mathbf{\Phi}\in\mathbb{R}^{m\times c}$, which yields the predicted class probabilities.
These predictions are compared against the ground-truth labels using a cross-entropy loss, which is minimized through gradient-based optimization.
%Note that a subset of the subsequent experiments is conducted within this framework, including ablation studies and parameter analyses.

\section{Complexity Analysis}\label{sec:com_ana}
The dominant computational cost of THR lies in: 
1) Torque computation and graph rewiring.
For each order $l$, we compute torque values only on the edges in $\mathcal{A}^{(l)}$, costing $\mathcal{O}(|\mathcal{A}^{(l)}|)$, and then sort these values in $\mathcal{O}(|\mathcal{A}^{(l)}|\log|\mathcal{A}^{(l)}|)$.
When adding edges, if the candidate set size is $B$, the combined probability calculation and sorting cost is $\mathcal{O}(B+B\log B)$.
2) Message passing on the rewired graph $\mathcal{A}^{(l)}$.
For the input layer with parameter $\mathbf{\Theta}\in \mathbb{R}^{d\times m}$ on $\mathbf{X}\in \mathbb{R}^{N\times d}$, it costs $\mathcal{O}(Ndm)$.
Aggregation over $\mathcal{A}^{(l)}$ then costs $\mathcal{O}(m|\mathcal{A}^{(l)}|)$ per layer. The output layer with $\mathbf{\Phi}\in \mathbb{R}^{m\times c}$ requires $\mathcal{O}(Nmc)$.
Putting these together for an $L$-layer network and assuming $B\ll |\mathcal{A}^{(l)}|$ for all $l$, the overall complexity is $\mathcal{O}(Ndm+\sum_{l=1}^L|\mathcal{A}^{(l)}|\log |\mathcal{A}^{(l)}|)$, which is slightly higher than that of standard methods with $\mathcal{O}(Ndm + m|\mathcal{A}^{(l)}|)$.

\section{Experiments}
% In the section, we construct a series of experiments to assess the proposed TorqueGNN.
% Our model is implemented in PyTorch on a workstation with AMD Ryzen 9 5900X CPU (3.70GHz), 64GB RAM and RTX 3090GPU (24GB caches). Our code is available at \url{https://anonymous.4open.science/r/TorqueGNN-F60C/README.md}.

\textbf{Datasets.} 
We evaluate our method on eleven standard node classification benchmarks, which include six heterophilic datasets: Texas, Wisconsin, Cornell, Actor, Penn94 and Flickr; five homophilous graphs Citeseer, Cora, Pubmed, Tolokers and Questions. 
Among them, Tolokers, Questions, Penn94 and Flickr are large-scale datasets.
The statistics for these datasets are summarized in Table~\ref{tab:datasets}, with further details provided in Appendix~\ref{sec:datasets}.
\begin{wraptable}{r}{0.55\textwidth}
\centering
\caption{Benchmark dataset statistics.}\label{tab:datasets}
\resizebox{0.55\textwidth}{!}{
	\begin{tabular}{cccccc}
		\toprule
		Datasets  & Edge Hom. & \#Nodes & \#Edges & \#Classes & \#Features \\
		\midrule
		Texas     & 0.11      & 183     & 295     & 5         & 1,703      \\
		Wisconsin & 0.21      & 251     & 466     & 5         & 1,703      \\
		Cornell   & 0.30       & 183     & 280     & 5         & 1,703      \\
		Actor     & 0.22      & 7,600   & 26,752  & 5         & 931        \\
		Citeseer  & 0.74      & 3,327   & 4,676   & 7         & 3,703      \\
		Cora      & 0.81      & 2,708   & 5,278   & 6         & 1,433      \\
		Pubmed    & 0.80       & 19,717  & 44,327  & 3         & 500      \\
		Tolokers &  0.60      & 11,758  & 51,900  & 2         & 10  \\
		Questions & 0.84       & 48,921  & 153,540  & 2         & 301  \\
		Penn94 & 0.47       & 41,554  & 1,362,229  & 2         & 4,814 \\
		Flickr & 0.32       & 89,250  & 2,724,458  & 7         & 500 \\
		\bottomrule
	\end{tabular}
}
\end{wraptable}

\textbf{Baselines.}
THR is a plug-in module that can be integrated into various state-of-the-art GNNs.
To evaluate the improvements offered by THR for GNNs, we select three representative models for experimentation, including two models designed for homophilous graphs: the vanilla GCN \cite{GCN} and the deep-based APPNP \cite{APPNP}, and GPRGNN \cite{GPRGNN} that is designed for heterophilous graphs.

To evaluate the effectiveness of THR in comparison to other graph rewiring techniques, we select five superior methods, including: First-order Dpectral Rewiring (FoSR) \cite{fosr}, Batch Ollivier-Ricci Flow (BORF) \cite{borf}, Stochastic Jost and Liu Curvature Rewiring (SJLR) \cite{SJLR}, Deep Heterophily Graph Rewiring (DHGR) \cite{DHGR} and randomly edge removal (DropEdge).
Here, we adopt layer-wise DropEdge (Dropedge-L), as proposed by \cite{rong2019dropedge}, to ensure a fair comparison with the hierarchical structure of THR. Further details on all methods are provided in Appendix \ref{sec:methods}.

\textbf{Setups.}
We report node classification accuracy (ACC), defined as the proportion of correctly predicted labels. 
For all benchmark datasets, models are trained using the Adam optimizer.
% We compare all rewiring methods on GCN designed for homophilous graphs and GPRGNN designed for heterophilous benchmarks.
% APPNP is used to ablation study and parameter sensitity.
\textit{Competitors are performed based on their respective source code.}
Detailed hyperparameters and environment configurations for THR are provided in Appendix \ref{sec:hyper}.
Following prior work \cite{GGCN, pei2020geom}, we adopt the data split strategy for all methods: 48\% of the nodes are used for training, 32\% for validation, and the remaining 20\% for testing. %\footnote{As shown in \cite{GGCN}, the actual data split used in \cite{pei2020geom} is 48\%/32\%/20\%.}.
%For the specific hyperparameter of THR, the number of candidate edge $t$ is selected from $\{1, 2, 4, 6, 8, 10\}$.
Each experiment is conducted over 10 runs with different random splits, and the results are reported as the mean and standard deviation.
\begin{table}[!htbp]
\centering
\caption{Node classification results on benchmark datasets with GCN and GPRGNN as the backbone models: Mean ACC \% (Standard Deviation \%). The first- and second-best accuracies are highlighted in {\color[HTML]{FF0000}\textbf{red}} and {\color[HTML]{00B050}\textbf{green}}, respectively.}\label{tab:performance}
\resizebox{1\textwidth}{!}{
	\begin{tabular}{cccccccc}
		\toprule
		Methods/Datasets & Citeseer              & Cora                  & Pubmed                & Texas                 & Wisconsin               & Actor  & Cornell                 \\
		\midrule
		GCN              & 75.52 (2.19)          & 86.96 (1.27)          & 86.43 (0.38)          & 58.61 (7.18)          & 52.60 (8.72)          & 30.15 (1.03) & 57.50 (4.66)         \\
		FoSR             & 78.03 (1.45)          & {\color[HTML]{FF0000}\textbf{87.00 (1.21)}} & 86.34 (0.31)          & {\color[HTML]{00B050}\textbf{74.70 (6.23)}}          & 65.58 (4.89)          & 30.16 (1.03)    & 54.59 (5.01)      \\
		BROF             & 78.45 (1.52)          & 86.86 (1.35)          & 86.42 (0.38)          & 74.51 (6.26)          & 65.59 (4.52)          & 30.20 (1.17)  & {\color[HTML]{FF0000}\textbf{60.27 (3.64)}}         \\
		SJLR             & 77.87 (1.81)          & 86.60 (1.64)          & {\color[HTML]{00B050}\textbf{86.52 (1.73)}}          & 60.14 (0.89)          & 55.16 (0.95)          & 30.80 (1.34)  & 58.11 (6.86)        \\
		DHGR             & {\color[HTML]{00B050}\textbf{78.68 (1.51)}}          & 86.61 (1.73)          & 86.40 (0.38)          & 60.20 (6.39)          & {\color[HTML]{00B050}\textbf{66.07 (12.51)}}         & {\color[HTML]{FF0000}\textbf{34.39 (0.99)}} & 58.68 (5.01) \\
		DropEdge-L       & 74.93 (1.85)          & 86.62 (1.23)          & 83.07 (2.58)          & 62.74 (8.32)          & 58.82 (8.24)          & 32.97 (0.92) &  54.32 (3.72)        \\
		THR              & {\color[HTML]{FF0000}\textbf{80.43 (1.52)}} & {\color[HTML]{00B050}\textbf{86.97 (1.19)}}          & {\color[HTML]{FF0000}\textbf{87.21 (0.45)}} & {\color[HTML]{FF0000}\textbf{76.27 (4.67)}} & {\color[HTML]{FF0000}\textbf{68.09 (2.71)}} & {\color[HTML]{00B050}\textbf{33.20 (0.90)}}  & {\color[HTML]{00B050}\textbf{58.91 (9.11)}}        \\
		\midrule
		\midrule
		GPRGNN           & 77.37 (1.83)          & 87.34 (1.14)          & 87.21 (0.43)          & 89.22 (5.56)          & 87.94 (5.29)          & 37.27 (1.16)  & 80.27 (6.63)        \\
		FoSR             & 77.37 (1.83)          & {\color[HTML]{00B050}\textbf{87.52 (1.63)}}          & 87.22 (0.46)          & 90.20 (5.04)          & {\color[HTML]{00B050}\textbf{89.85 (3.45)}}          & 37.25 (1.19) & 84.05 (7.88)         \\
		BORF             & {\color[HTML]{00B050}\textbf{78.77 (1.67)}}          & 87.49 (1.24)          & 87.17 (0.39)          & {\color[HTML]{00B050}\textbf{91.16 (5.15)}}          & 89.11 (4.32)          & 37.52 (1.06) & {\color[HTML]{00B050}\textbf{85.49 (4.83)}}         \\
		SJLR             & 78.38 (1.49)          & 86.97 (1.63)          & {\color[HTML]{00B050}\textbf{88.11 (0.41)}}          & 90.00 (2.83)          & 89.26 (6.38)          & 34.87 (1.69) & 81.62 (9.35)         \\
		DHGR             & 77.77 (2.06)          & 87.19 (1.39)          & 87.69 (0.47)          & 89.02 (4.31)          & 86.03 (6.32)          & 35.20 (1.20)      & 84.31 (4.56)    \\
		DropEdge-L       & 78.73 (1.91)          & 86.91 (1.07)          & 87.50 (0.48)          & 90.17 (3.06)          & 87.79 (6.28)          & {\color[HTML]{00B050}\textbf{37.77 (1.16)}}  & 84.05 (9.00)        \\
		THR              & {\color[HTML]{FF0000}\textbf{79.15 (1.69)}} & {\color[HTML]{FF0000}\textbf{87.60 (1.15)}} & {\color[HTML]{FF0000}\textbf{88.28 (0.52)}} & {\color[HTML]{FF0000}\textbf{91.96 (3.76)}} & {\color[HTML]{FF0000}\textbf{91.91 (4.75)}} & {\color[HTML]{FF0000}\textbf{38.00 (0.56)}} & {\color[HTML]{FF0000}\textbf{86.22 (5.19)}} \\
		\bottomrule
	\end{tabular}
}
\end{table}

\textbf{Node Classification Results.}
Table \ref{tab:performance} presents the test-set accuracy gains achieved by various rewiring approaches on GCN and GPRGNN across seven benchmark datasets. Several key insights can be drawn:
1) Compared to the baselines, all rewiring methods show performance improvements on most datasets, with particularly notable gains on heterophilous graphs.
2) In all datasets, the proposed THR ranks among the top two performers, achieving the highest accuracy gain on the majority of benchmarks.
3) Although FoSR, BORF, and DHGR also exhibit strong performance on certain datasets, their gains are only marginally higher than those of THR. Overall, THR outperforms these methods and delivers the best results in all cases when GPRGNN is used as the downstream model.
4) DropEdge-L, which is also based on hierarchical graph rewiring, outperforms other rewiring methods on some datasets (e.g., Texas and Actor), validating the effectiveness of the hierarchical strategy.
Although DropEdge shows performance improvements on certain datasets, its inherent randomness negatively impacts the model’s performance, resulting in lower performance than the baseline in some cases, e.g., Citeseer. 
This further validates the effectiveness of the proposed torque-driven hierarchical approach.

\begin{wraptable}{l}{0.6\textwidth}
\centering
\caption{Node classification results on \textbf{large-scale} datasets: Mean ACC \% (ROC
	AUC for imbalanced Questions and Tolokers) (Standard Deviation \%), where the optimal and suboptimal results are highlighted in {\color[HTML]{FF0000}\textbf{red}} and {\color[HTML]{00B050}\textbf{green}}, respectively. OoM means that the model suffers from the out-of-memory error.}
\resizebox{0.6\textwidth}{!}{
	\begin{tabular}{cccccc}
		\toprule
		Methods/Datasets & Questions & Tolokers & Penn94 & Flickr \\
		\midrule
		GCN & {\color[HTML]{00B050}\textbf{75.26 (0.84)}} & 83.79 (0.74) & 80.18 (0.36) & 57.48 (7.85) \\
		FoSR & 75.19 (0.71) & {\color[HTML]{00B050}\textbf{84.14 (0.99)}} & 80.19 (0.35) & 58.03 (6.75) \\
		BROF & 75.15 (0.84) & MemoryError & OoM & OoM \\
		SJLR & 72.07 (6.12) & 76.14 (1.14) & {\color[HTML]{00B050}\textbf{80.20 (0.28)}} & {\color[HTML]{00B050}\textbf{64.49 (2.82)}} \\
		DHGR & OoM & 76.45 (12.16) & OoM & OoM \\
		DropEdge-L & 74.06 (1.11) & 84.00 (0.65) & 62.27 (0.35) & 59.29 (2.26) \\
		THR & {\color[HTML]{FF0000}\textbf{75.92 (1.09)}}& {\color[HTML]{FF0000}\textbf{84.43 (0.88)}}  & {\color[HTML]{FF0000}\textbf{80.32 (0.23)}} & {\color[HTML]{FF0000}\textbf{68.29 (0.78)}} \\
		\midrule
		GPRGNN & 72.89 (1.42) & 71.99 (0.93) & 84.18 (0.30) & 48.90 (5.81) \\
		FoSR & 72.91 (1.43) & {\color[HTML]{00B050}\textbf{71.99 (0.93)}} & {\color[HTML]{00B050}\textbf{84.22 (0.29)}} & 49. 16 (6.49) \\
		BORF & {\color[HTML]{00B050}\textbf{72.99 (1.44)}} & MemoryError & OoM & OoM \\
		SJLR & 72.27 (1.24) & 69.46 (1.07) & 83.89 (0.20) & 61.61 (3.22) \\
		DHGR & OoM & 70.96 (1.14) & OoM & OoM \\
		DropEdge-L & 72.07 (1.35) & 71.98 (1.09) & 83.67 (0.44) & {\color[HTML]{00B050}\textbf{63.42 (3.26)}} \\
		THR & {\color[HTML]{FF0000}\textbf{73.41 (0.98)}} & {\color[HTML]{FF0000}\textbf{72.05 (1.24)}} & {\color[HTML]{FF0000}\textbf{84.45 (0.29)}} & {\color[HTML]{FF0000}\textbf{65.29 (3.30)}} \\
		\bottomrule
	\end{tabular}\label{tab:large_performance}
}
\end{wraptable}
\textbf{Results on Larger Graphs.}
Scalability of rewiring techniques on large graphs is crucial, particularly for end-to-end methods that dynamically add and remove edges during training.
In THR, the primary computational cost arises from computing torques and the corresponding gaps, which incurs a complexity of $\mathcal{O}(|\mathcal{A}^{(l)}|\log |\mathcal{A}^{(l)}|)$ (see Section \ref{sec:com_ana} for details).
Despite this overhead, THR remains computationally feasible for large graphs.
Table \ref{tab:large_performance} compares several rewiring schemes on larger datasets, with THR consistently outperforming all alternatives.
Notably, with the exception of the Flickr dataset, all rewiring methods show only marginal improvements, and in some cases, even lead to a decline in performance.
This may be attributed to the fact that the raw graphs of these datasets already contain sufficient structural information, and the rewiring methods introduce only minor modifications, or in some cases, may even result in the loss of critical semantics, thus negatively impacting classification performance.

\begin{wrapfigure}{r}{0.6\textwidth}
\centering
\begin{minipage}{0.6\textwidth}
	\centering
	\includegraphics[width=\linewidth]{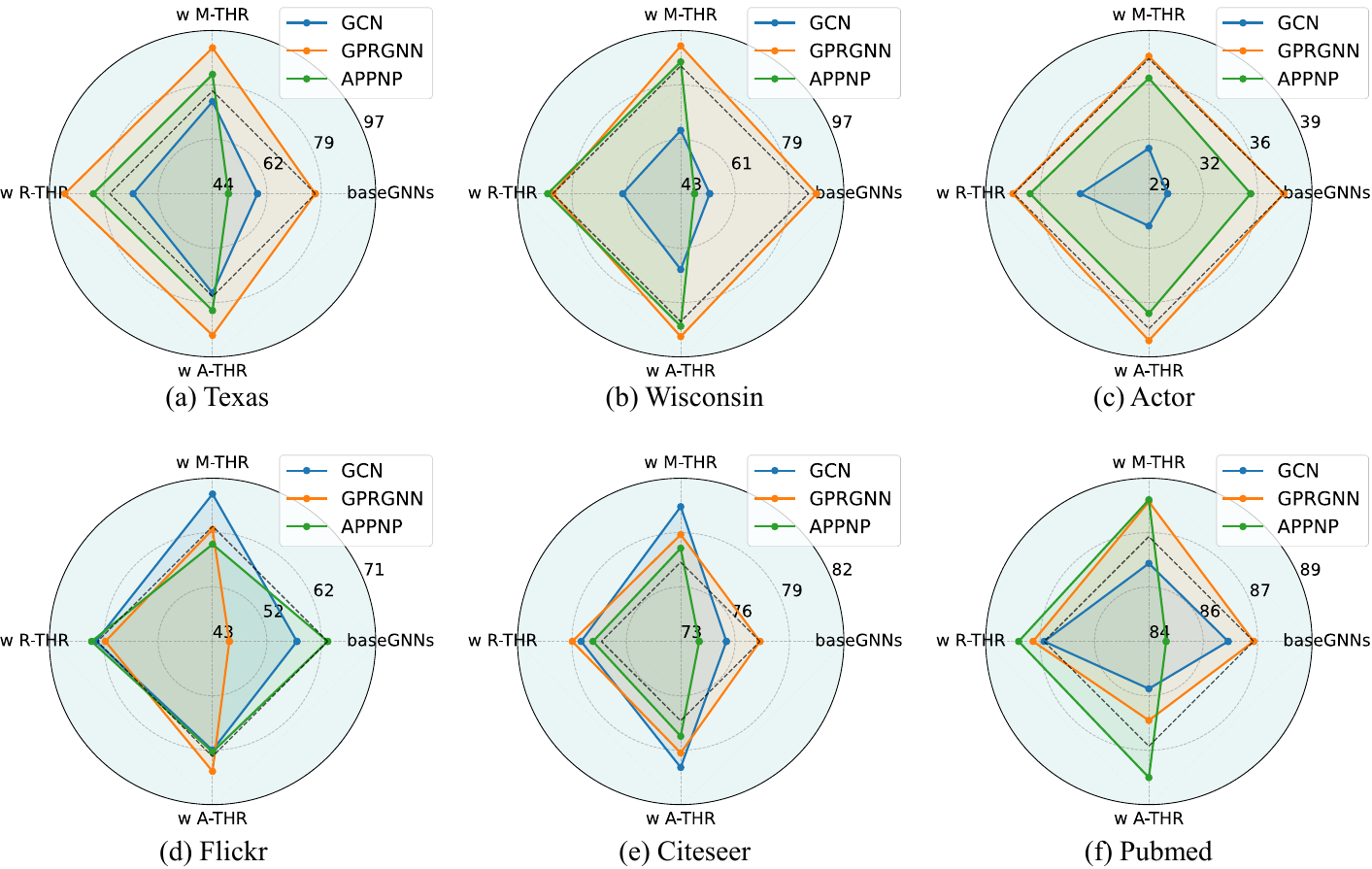}
	% \caption{(a) Texas}
\end{minipage}
\caption{Ablation study: Performance comparison of GCN, GPRGNN, and APPNP with various THR variants across six datasets.}
\label{fig:ablation}
\end{wrapfigure}
\textbf{Ablation Study.}
We conduct an ablation study to assess the impact of edge removal and addition operations in THR, using GCN, GPRGNN, and APPNP as backbone models.
THR has three variants: edge-addition THR (A-THR), edge-removal THR (R-THR), and mixed THR (M-THR).
As illustrated in Figure \ref{fig:ablation}, most rewiring variants significantly outperform their base GNNs.
However, on the Wisconsin dataset, GPRGNN slightly surpasses GPRGNN w R-THR, likely because GPRGNN effectively allocates sign edges to distinguish class information, while R-THR removes heterophilic links, inadvertently causing the model to lose some discriminative features.
On the Flickr dataset, R-THR improves performance for all models, while A-THR and M-THR degrade the performance of APPNP. 
Similarly, on PubMed, both A-THR and M-THR reduce the performance of GCN.
These results suggest that for these graphs, excessive edge addition leads to information interference and confusion of node features, while GPRGNN mitigates this effect by utilizing a sign edge strategy.
In conclusion, the THR strategy enhances model performance, but its effectiveness varies across datasets with different characteristics.

Moreover, to investigate the significance of the proposed torque, which integrates feature distance and homophily ratio disparity from a physical perspective, we evaluate THR and its variants based on the edge-removal strategy.
THR$_{\text{dis.}}$ refers to the method of removing edges based on the distance metric between node pairs.
THR$_{\text{torque w/o homo.}}$ leverages the torque  without considering the disparity in homophily ratio to drop edges.
THR$_{\text{w/o H}}$ denotes the version of THR without hierarchical rewiring, where all layers share the same graph. 
Table \ref{tab:abla} displays the ablation results, showing that the node classification accuracy of variants that do not use the proposed torque decreases across all heterophilous datasets.
Moreover, on the Cornell and Flickr datasets,  THR$_{\text{w/o H}}$ outperforms THR, suggesting that layer-wise rewiring may excessively complicate their graph structures, thereby hindering the propagation of effective information.
In summary, both THR and THR$_{\text{w/o H}}$ rely on the proposed torque for graph rewiring and both rank in the top two across all datasets.
This validates that THR effectively models heterophilous graphs by integrating the distance and homophily ratio disparities between node pairs from a physical perspective.
%Ablation study of significant metrics is included in Appendix \ref{sec:more_exper}.
\begin{table}[!htbp]
\centering
\caption{Ablation study: A comparison of THR and its variants by removing specific components. The optimal and suboptimal results are highlighted in bold and underlined, respectively.}\label{tab:abla}
\resizebox{1\textwidth}{!}{
	\begin{tabular}{ccccccc}
		\toprule
		Datasets      & Texas                 & Wisconsin             & Cornell               & Actor                 & Penn94                & Flickr                \\
		\midrule
		THR$_{\text{dis.}}$      & 70.39 (9.62)          & 74.56 (7.21)          & 76.22 (8.53)          & 35.80 (1.27)          & 74.98 (0.55)          & 61.51 (4.32) \\
		THR$_{\text{torque w/o homo.}}$ & 67.84 (10.96)         & 72.94 (8.71)          & 76.03 (8.40)          & 35.57 (1.31)          & 75.94 (0.57)          & 61.23 (4.74)          \\
		THR$_{\text{w/o H}}$ & \underline{70.98 (8.12)}         &  \underline{75.29 (4.45)}          & \textbf{78.65 (6.78)}          & \underline{35.96 (1.33)}         & \underline{76.14 (0.63)}          & \textbf{64.13 (2.27)} \\
		THR        & \textbf{72.01 (6.13)} & \textbf{75.89 (3.46)} & \underline{77.30 (7.17)} & \textbf{36.28 (1.15)} & \textbf{76.21 (0.47)} & \underline{63.28 (1.56)}         \\
		\bottomrule
	\end{tabular}
}
\end{table}

\begin{figure}[!htbp]
\centering
\begin{minipage}{1\textwidth}
	\centering
	\includegraphics[width=\linewidth]{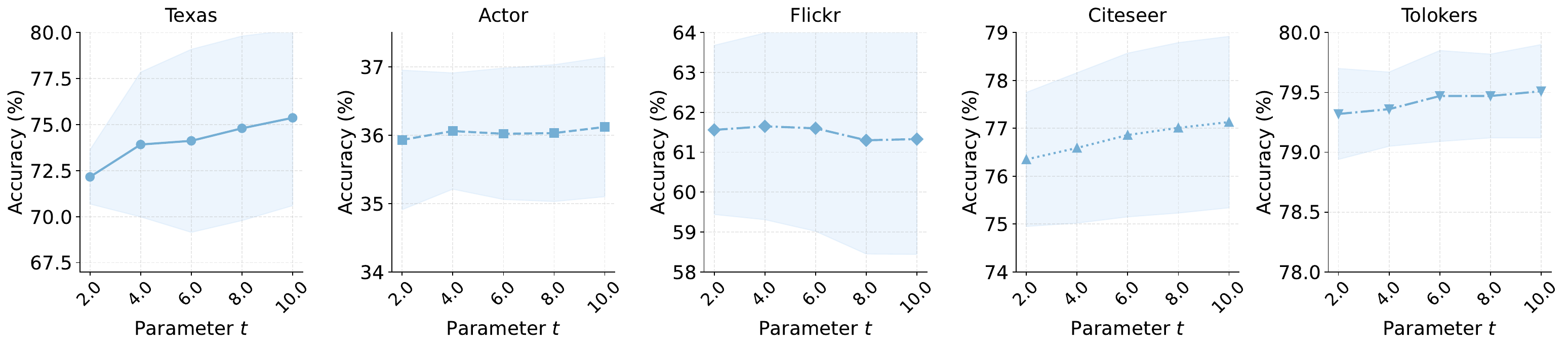}
\end{minipage}
\caption{Parameter sensitivity: Performance curves on five datasets as the number of candidate edges $t$ varies from 2 to 10.}\label{fig:t}
\end{figure}

\textbf{Parameter Analysis.}
Since the edge-removal procedure automatically determines the cutoff $\mathcal{K}$, we investigate the main hyperparameter $t$ of THR, which defines the number of candidate edges for addition.
As shown in Figure \ref{fig:t}, we present the performance curves for varying $t$ values in $\{2, 4, 6, 8, 10\}$ across five datasets.
On both homophilous and heterophilous datasets, accuracy increases as $t$ grows, demonstrating that the proposed edge-addition scheme aids the model in capturing global information.
However, this does not imply that adding more edges is always beneficial.
For instance, on the Flickr dataset, performance decreases when $t=8$, as excessive edge addition may introduce noise, as highlighted in the ablation study.
Sensitivity analysis of other hyperparameters is presented in Appendix \ref{sec:more_exper}.

\section{Conclusion}
In summary, we proposed a \textbf{Torque}‑driven \textbf{h}ierarchical \textbf{r}ewiring strategy (THR), which dynamically refined the graph structures to enhance representation learning on heterophilous and homophilous graphs.
By introducing an interference‑aware torque metric, the product of the displacement vector and the feature vector weighted by the homophily ratio disparity, THR automatically removed undesirable connections and introduced beneficial ones during message passing.
This hierarchical rewiring yielded interference‑resilient, importance‑aware propagation tailored to each layer’s receptive field.
Extensive evaluations across homophilous and heterophilous benchmark datasets demonstrated that THR consistently obtained the performance gains and outperformed other rewiring methods.
%We state the limitations of our model in Appendix \ref{sec:limi}.

%\bibliography{iclr2026_conference}
%\bibliographystyle{iclr2026_conference}

\appendix
\section{Appendix}
\section{Algorithm}\label{sec:alg}
Algorithm \ref{algorithm} outlines the complete workflow of APPNP with THR. 
\begin{algorithm}[!htbp]
\caption{GNN with THR}\label{algorithm}
\KwIn{Node features $\{\mathbf{x}_i\in \mathbb{R}^{ d}\}_{i=1}^N$, candidate edge set $\mathcal{T}$, ground truth matrix $\mathbf{Y}$, the number of layers $L$, hyperparameters $t$ and $\alpha$.}
\KwOut{The predicted class label.}
Initialize network parameters $\mathbf{\Theta}, \mathbf{\Phi}$;\\
$\mathbf{h}_i^{(0)}=\text{ReLU}(\mathbf{x}_i\mathbf{\Theta})$\;
\For {$l = 1 \rightarrow L$}
{
	$\vartriangleright$ \textbf{Forward Propagation}\\
	Compute pairwise distance $D_{\langle i, j \rangle}^{(l)}$ and homophily disparity $E_{\langle i, j\rangle}$ with Eqs. \ref{eq:distance} and \ref{eq:edge_energy}; \\
	Compute the $l$th order torques with Eq. \ref{eq:graphTorque_value} and sort them. 
	// Torque computation \\
	Gain the largest torque gap $\mathcal{K}$ with Eqs. \ref{eq:gap}-\ref{eq:weight};\\
	Remove the top $\mathcal{K}$ edges to gain $\mathcal{A}^{(l)^*}$. 
	// Removing undesirable edges \\
	Compute the sampling probability of candidate edges with Eq. \ref{eq:gumbel_softmax}; \\
	Add beneficial candidate edges to form the refined propagation matrix $\mathcal{A}^{(l)}$. 
	// Adding desirable connections \\
	Update node representation $\mathbf{h}_{i}^{(l)}$ with Eq. \ref{eq:intra_rep_updat}. // Message passing \\
	$\vartriangleright$ \textbf{Backward Propagation}\\
	Classifier $f(\cdot)$ $\longleftarrow$ LocalUpdating($\mathbf{x}_i, \{\mathcal{A}^{(l)}\}_{l=1}^L$) with the cross-entropy loss // Standard training \\
}
Obtain $\mathbf{\hat{y}}_i = \text{Softmax}\left(\mathbf{h}^{(L)}_i\mathbf{\Phi}\right)$\;
\Return{\rm{The predicted class label of the $i$-th node is given by} $\arg\max{\hat{\mathbf{y}}_{i}}$.}
\end{algorithm}

% \section{Complexity Analysis}\label{sec:com_ana}
% The dominant computational cost of TorqueGNN lies in: %1) High-order neighbor extraction. 
% %Given a sparse adjacency matrix $\mathbf{A}$ with $K$ entries, computing its $l$-th power to identify all $l$-hop neighbors incurs $\mathcal{O}(l\frac{K^2}{N})$.
% 1) Torque computation and graph rewiring.
% %Considering rewired propagation matrix $\mathcal{A}^{(l)}$, 
% For each order $l$, we compute torque values only on the edges in $\mathcal{A}^{(l)}$, costing $\mathcal{O}(|\mathcal{A}^{(l)}|)$, and then sort these values in $\mathcal{O}(|\mathcal{A}^{(l)}|\log|\mathcal{A}^{(l)}|)$.
% When adding edges, if the candidate set size is $B$, the combined probability calculation and sorting cost is $\mathcal{O}(B+B\log B)$.
% 2) Message passing on the rewired graph $\mathcal{A}^{(l)}$.
% Initial representations $\mathbf{H}^{(0)}$ are obtained by a fully connected layer with parameter $\mathbf{\Theta}\in \mathbb{R}^{d\times m}$ on $\mathbf{X}\in \mathbb{R}^{N\times d}$, at cost $\mathcal{O}(Ndm)$.
% Aggregation over $\mathcal{A}^{(l)}$ then costs $\mathcal{O}(m|\mathcal{A}^{(l)}|)$ per layer. The output layer with $\mathbf{\Phi}\in \mathbb{R}^{m\times c}$ requires $\mathcal{O}(Nmc)$.
% Putting these together for an $L$-layer network and assuming $B\ll |\mathbf{A}^{l}|$ for all $l$, the overall complexity is $\mathcal{O}(LNdm+\sum_{l=1}^L|\mathcal{A}^{(l)}|\log |\mathcal{A}^{(l)}|)$.

\section{More Experimental Results}
\subsection{Configures}\label{sec:config}
We construct a series of experiments to assess the proposed TorqueGNN.
Our model is implemented in PyTorch on a workstation with AMD Ryzen 9 5900X CPU (3.70GHz), 64GB RAM and RTX 3090GPU (24GB caches). Our code is available at \url{https://anonymous.4open.science/r/TorqueGNN-F60C/README.md}.

\subsection{Datasets}\label{sec:datasets}
\begin{itemize}
\item Homophilous Datasets. Citeseer, Cora and Pubmed are three citation networks, and they are published in \cite{sen2008collective}.
Specifically, 
\begin{itemize}
	\item \textbf{Citeseer}  comprises 3,327 publications classified into six categories, with each paper encoded by a 3,703-dimensional binary word‐presence vector.
	\item \textbf{Cora} consists of 2,708 scientific publications classified into seven research topics. Each paper is represented by a 1,433-dimensional binary feature vector indicating the presence of specific word
	\item \textbf{Pubmed} is a larger citation network of 19,717 diabetes-related articles labeled among three classes. 
	Papers are described by 500-dimensional term frequency–inverse document frequency feature vectors, and citation edges capture scholarly references.
	\item \textbf{Tolokers} \cite{platonov2023critical} is built from the Toloka crowdsourcing platform, comprising 11,758 nodes and 519,000  edges that link workers who collaborated on the same task. 
	Each node carries a 10-dimensional feature vector and is assigned one of two labels based on whether the worker was banned.
	\item \textbf{Questions} \cite{platonov2023critical} is an interaction graph of users on the Yandex Q question-answering platform, comprising 48,921 nodes and 153,540 edges that link users who interacted on the same question. Each node carries a 301-dimensional feature vector and a binary label for node classification.
\end{itemize}
\item Heterophilous Datasets
\begin{itemize}
	\item \textbf{Texas}, \textbf{Wisconsin}, \textbf{Cornell} are WebKB datasets used in \cite{pei2020geom}, where nodes correspond to individual web pages and edges correspond to the hyperlinks between them. Every node is described by a bag-of-words feature vector extracted from its page content, and each page has been manually labeled into one of five categories. 
	\item \textbf{Actor} \cite{tang2009social} is the actor-only induced subgraph of a film–director–actor–writer network on Wikipedia, where each node represents an actor and an undirected edge connects two actors if they co-occur on the same Wikipedia page.
	\item \textbf{Penn94} \cite{lim2021large} is a subgraph of the Facebook100 dataset featuring 41,554 university students as nodes, connected by 1,362,229 undirected friendship edges.
	Each node is described by a five-dimensional feature vector and labeled by the gender of the students.
	\item \textbf{Flickr} \cite{flickr} is an undirected graph originated from NUS-wide, including 89,250 nodes and 2,724,458 edges.
	Each node is an image with 500-dimensional bag-of-word features and each edge links two images sharing some common properties.
\end{itemize}
\end{itemize}

\subsection{Baselines}\label{sec:methods}
\subsubsection{GNNs for Homophilous and Heterophilous Graphs}
\textbf{GCN} generalize convolutional neural networks to graph-structured data by iteratively aggregating feature information from each node’s local neighborhood, 
\begin{equation}\label{met:gcn}
\mathbf{h}^{(l)}_i = \sigma(\widetilde{\mathbf{A}}\mathbf{h}^{(l-1)}_{i}\mathbf{W}^{(l)}),
\end{equation}
where $\mathbf{W}$ is the learnable parameter matrix.

\textbf{APPNP} first achieves the feature transformation by:
\begin{equation}\label{me:appnp}
\mathbf{H}^{(0)} = \mathbf{X}\mathbf{W},
\end{equation}
and then propagating message via a Personalized PageRank scheme:
\begin{equation}
\mathbf{H}^{(l)} = (1-\alpha)\mathbf{P}\mathbf{H}^{(l-1)} + \alpha \mathbf{H}^{(0)}.
\end{equation}
Here, $\mathbf{P}=\mathbf{D}^{-1/2}\mathbf{A}\mathbf{D}^{-1/2}$ is the symmetrically normalized adjacency matrix and $\alpha$ is a trade-off hyperparameter.

\textbf{GPR-GNN} generalizes personalized PageRank by treating each hop’s contribution as a learnable parameter:
\begin{equation}
\mathbf{H} = \sum_{l=1}^L\gamma^l\mathbf{P}\mathbf{H}^{(0)}, \mathbf{H}^{(0)}=\mathbf{H}\mathbf{W},
\end{equation}
where $\gamma^l\mathbf{P}$ measures the propagation coefficient for the connection between nodes $v_i$ and $v_j$.

\subsubsection{Rewiring Strategies}
\textbf{DropEdge} randomly remove edges at each training epoch to act as both data augmentation and message‑passing reduction, which is used to mitigate over-fitting and over-smooting problems.

\textbf{FoSR} is a preprocessing method, which aim to address the oversquashing issue by improving the graph connectivity.  It adds edges by exploring the first order change in the spectral gap. 

\textbf{BORF} uses the Ollivier-Ricci curvature to rewire graph, where minimally curved edges causing the information bottlenecks should add connections and maximally curved edges leading to over-smoothing should be removed.

\textbf{SJLR} combines the Jost–Liu Curvature of each edge with the embedding similarity between its incident nodes, and uses the weighted score as the probability for edge removal or addition.

\textbf{DHGR} compares the neighborhood feature distribution and neighborhood label distribution between node pairs; edges connecting nodes with low similarity (heterophilous) are pruned, while edges between highly similar (homophilous) nodes are added.

\textbf{DHGR vs. THR}. Although both methods essentially assess edge homophily or heterophily through feature and label differences, they follow distinct methodological lines.
DHGR is a preprocessing approach that aggregates neighborhood features and derives local label distributions from pseudo-labels produced by a pre-trained model, a heuristic design without explicit theoretical grounding.
By contrast, THR operates within the optimizing model, contrasting node representations and quantifying their homophily ratio disparity, thereby aligning with prior theoretical proofs and offering a more principled formulation.

\subsection{Hyperparameters}\label{sec:hyper}
In the subsection, we list the detailed hyperparameters used for the experiments and they are also provided in code.
The hyperparameters can be found in Tables \ref{tab:hyper_gcn}-\ref{tab:hyper_appnp}.

\begin{table}[!htbp]
\centering
\caption{Hyperparameters of THR on GCN across 11 datasets.}\label{tab:hyper_gcn}
\begin{tabular}{ccccccccc}
	\toprule
	Datasets & Lr    & Wd     & 
	Dropout & L & t  & epochs & Normalize Data & Hidden Size \\
	\midrule
	Texas     & 0.05  & 0.0005 & 0.5     & 2 & 5  & 10000  & Yes             & 32         \\
	Wisconsin & 0.05  & 0.0005 & 0.5     & 2 & 5  & 10000  & Yes             & 32         \\
	Cornell   & 0.05  & 0.0005 & 0.5     & 2 & 5  & 10000  & Yes             & 32         \\
	Actor     & 0.01  & 0.0005 & 0.5     & 2 & 2  & 10000  & No              & 32         \\
	Citeseer  & 0.01  & 0.0005 & 0.5     & 2 & 10 & 10000  & Yes             & 32         \\
	Cora      & 0.01  & 0.05   & 0.5     & 2 & 2  & 10000  & No              & 32         \\
	Pubmed    & 0.01  & 0.0005 & 0.5     & 2 & 2  & 10000  & Yes             & 32         \\
	\midrule
	Tolokers  & 0.005 & 5e-8  & 0.2     & 2 & 1  & 10000  & No              & 32         \\
	Questions & 0.005 & 5e-8  & 0.2     & 5 & 1  & 10000  & No              & 32         \\
	Penn94    & 0.001 & 5e-8  & 0.5     & 2 & 1  & 10000  & No              & 32         \\
	Flickr    & 0.01  & 0.0005 & 0.5     & 2 & 1  & 10000  & No              & 32   \\
	\bottomrule
\end{tabular}
\end{table}

\begin{table}[!htbp]
\centering
\caption{Hyperparameters of THR on GPRGNN across 11 datasets.}\label{tab:hyper_gcn}
\begin{tabular}{cccccccccc}
	\toprule
	Datasets & Lr    & Wd     & Dropout & L & t  & epochs & \multicolumn{1}{c}{Normalize Data} & PPR & Hidden Size \\
	Texas       & 0.05  & 0.0005 & 0.5     & 2 & 5  & 10000  & Yes                                 & 1   & 32         \\
	Wisconsin   & 0.05  & 0.0005 & 0.5     & 2 & 5  & 10000  & Yes                                 & 1   & 32         \\
	Cornell     & 0.05  & 0.0005 & 0.5     & 2 & 5  & 10000  & Yes                                 & 0.9 & 32         \\
	Actor       & 0.01  & 5e-8  & 0.5     & 2 & 2  & 10000  & Yes                                 & 0.9 & 32         \\
	Citeseer    & 0.01  & 0.0005 & 0.5     & 2 & 10 & 10000  & Yes                                 & 0.1 & 32         \\
	Cora        & 0.01  & 0.0005 & 0.5     & 2 & 5  & 10000  & Yes                                 & 0.1 & 32         \\
	Pubmed      & 0.05  & 0.0005 & 0.5     & 2 & 2  & 10000  & Yes                                 & 0.2 & 32         \\
	Tolokers    & 0.005 & 5e-8  & 0.5     & 2 & 1  & 10000  & No                                  & 0.1 & 256        \\
	Questions   & 0.05 & 5e-8  & 0.5     & 2 & 1  & 10000  & No                                  & 0.1   & 32        \\
	Penn94      & 0.01  & 0.0001 & 0.5     & 2 & 1  & 10000  & No                                  & 0.1 & 32         \\
	Flickr      & 0.01  & 0.0005 & 0.5     & 2 & 1  & 10000  & No                                  & 0.1 & 32   \\
	\bottomrule
\end{tabular}
\end{table} 

\begin{table}[!htbp]
\centering
\caption{Hyperparameters of THR on APPNP across 11 datasets.}\label{tab:hyper_appnp}
\begin{tabular}{cccccccccc}
	\toprule
	Datasets  & Lr    & Wd       & Dropout & $\alpha$ & $L$ & $t$ & epochs & Normalize Data & Hidden Size \\
	\midrule
	Texas     & 0.001 & 0.0005   & 0.7     & 0.05   & 8  & 5 & 100  & No  & 512        \\
	Wisconsin & 0.001 & 0.5      & 0.5     & 0.05   & 4  & 5 & 100  & No  & 512        \\
	Cornell   & 0.001 & 0.05     & 0.7     & 0.5   & 8  & 2 & 100  & No  & 512        \\
	Actor     & 0.001 & 0.05     & 0.1     & 0.5   & 8  & 5 & 100  & No  & 512        \\
	Citeseer  & 0.001 & 0.05     & 0.4     & 0.5   & 8  & 10 & 100  & No  & 512        \\
	Cora      & 0.001 & 0.5        & 0.4     & 0.5   & 4  & 2 & 100  & No  & 512        \\
	Pubmed    & 0.001 & 5e-8        & 0.4     & 0.5   & 4 & 2 & 100  & No  & 512        \\
	\midrule
	Tolokers  & 0.001 & 5e-8 & 0.1     & 0.8   & 2  & 2 & 500  & No  & 512        \\
	Questions & 0.001 & 5e-8 & 0.1     & 0.8   & 2  & 2 & 500  & No  & 512        \\
	Penn94    & 0.001 & 5e-8 & 0.1     & 0.8   & 2  & 2 & 500  & No  & 512        \\
	Flickr    & 0.001 & 0.5 & 0.1     & 0.8   & 2  & 2 & 500 & No  & 512   \\
	\bottomrule
\end{tabular}
\end{table}

\subsection{Experiments}\label{sec:more_exper}
\textbf{Classification Results.} Table \ref{tab:per_appnp} shows the performance gains brought by APPNP with diverse rewiring methods.
We can observe that on most datasets, THR obtains the optimal performance, indicating its effectiveness.

\begin{table}[!htbp]
\centering
\caption{Node classification results on benchmark datasets with APPNP as the backbone models: Mean ACC \% (Standard Deviation \%). The first- and second-best accuracies are highlighted in bold and underlined, respectively.}\label{tab:per_appnp}
\resizebox{1\textwidth}{!}{
	\begin{tabular}{cccccccc}
		\toprule
		Methods/Datasets & APPNP        & FoSR         & BROF         & SJLR         & DHGR         & DropEdge-L   & THR          \\
		\midrule
		Texas            & 49.17±3.30   & 78.04 (3.70) & 73.53 (7.66) & \underline{81.57 (3.94)} & 78.04 (4.62) & 72.75 (4.68) & \textbf{84.12 (3.22)} \\
		Wisconsin          & 47.60±4.54   & 74.26 (4.47) & 75.00 (4.41) & \underline{83.53 (5.95)} & 73.82 (3.77) & 71.91 (4.85) & \textbf{87.79 (3.54)} \\
		Actor            & 35.24 (0.56) & 35.51 (1.42) & 35.35 (1.28) & 35.19 (1.13) & \underline{35.90 (1.16)} & 35.48 (1.12) & \textbf{36.34 (0.87)} \\
		Cornell          & 67.57 (5.54) & 67.57 (5.54) & 68.38 (7.46) & \underline{74.59 (5.16)} & 69.46 (8.80) & 69.19 (7.27) & \textbf{ 77.57 (8.02)} \\
		Penn94           & 76.53 (0.28) & 76.53 (0.28) & OoM & \underline{79.73 (0.24)} & 79.10 (0.40) &  76.13 (0.40) & \textbf{82.56 (0.43)} \\
		Flickr           & 57.26 (7.69) & 56.31 (6.99) & OoM          & 61.86 (5.17) & \underline{62.20 (1.10)} &  61.25 (3.30)            & \textbf{63.28 (1.56)} \\
		Citeseer         & 74.02 (0.38)   & 77.65 (1.55) & 77.65 (1.24) & 77.25 (1.35) & 76.89 (1.81) & \underline{77.69 (1.67)} & \textbf{78.74 (1.29)} \\
		Cora             & 85.89 (1.19) & 85.89 (1.19) & 85.19 (1.87) & \textbf{86.51 (1.59)} & 85.85 (1.79) & 85.54 (1.14) & \underline{86.35 (1.61)} \\
		Pubmed           & 87.19 (0.55) & 87.19 (0.55) & 87.16 (0.40) & \textbf{88.84 (0.40)} & \underline{88.47 (0.44)} & 87.66 (0.33) & 88.31 (0.49) \\
		Tolokers         & 75.11 (0.74) & 75.11 (0.74) & OoM          & \underline{ 78.46 (1.11)} & 75.33 (0.83) & 74.64 (1.06) & \textbf{79.29 (0.42)} \\
		\bottomrule
	\end{tabular}
}
\end{table}

\begin{figure}[!htbp]
\centering
\begin{minipage}{1\textwidth}
	\centering
	\includegraphics[width=\linewidth]{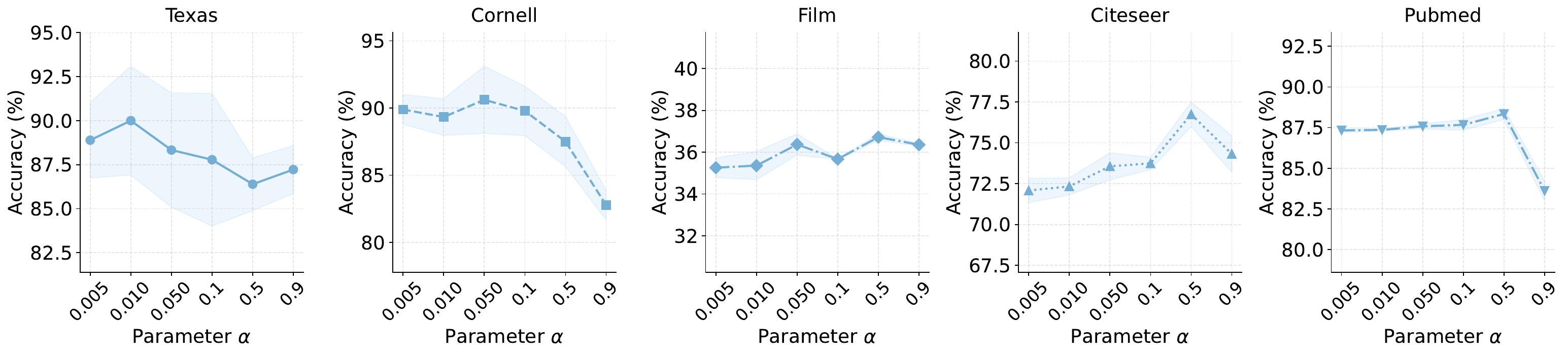}
\end{minipage}
\caption{Parameter sensitivity: Performance curves on five datasets with layers changing in $\{2, 4, 8, 16, 32\}$.}
\label{fig:alpha}
\end{figure}

\textbf{Parameter Sensitivity.} 
Although $\alpha$ balancing the contribution of the learned high-order representation and the original input features originates from APPNP, THR modifies the graph structure over which propagation occurs.	
To examine how signal diffusion changes with respect to $\alpha$ under the rewired graph, we perform a sensitivity analysis shown in Figure \ref{fig:alpha}, where a larger $\alpha$ increases the influence of the hidden representations.
%The figure examine the trade-off parameter $\alpha$, which balances the contribution of the learned hidden representation and the original input features, where a larger $\alpha$ increases the influence of the hidden representations.
We observe that 
smaller heterophilous graphs (e.g., Texas and Actor), optimal accuracy is achieved at low $\alpha=0.05$, implying that raw node features provide sufficient discriminative power. 
In contrast, on larger or homophilous graphs, better performance is observed when $\alpha=0.5$, reflecting the necessity of high-order hidden representations to capture more complex  community structures.
Moreover, for all datasets, the best results are gained at a larger $\alpha$, which demonstrates the effectiveness of excavating  deep features.

\begin{figure}[!htbp]
\centering
\begin{minipage}{1\textwidth}
	\centering
	\includegraphics[width=\linewidth]{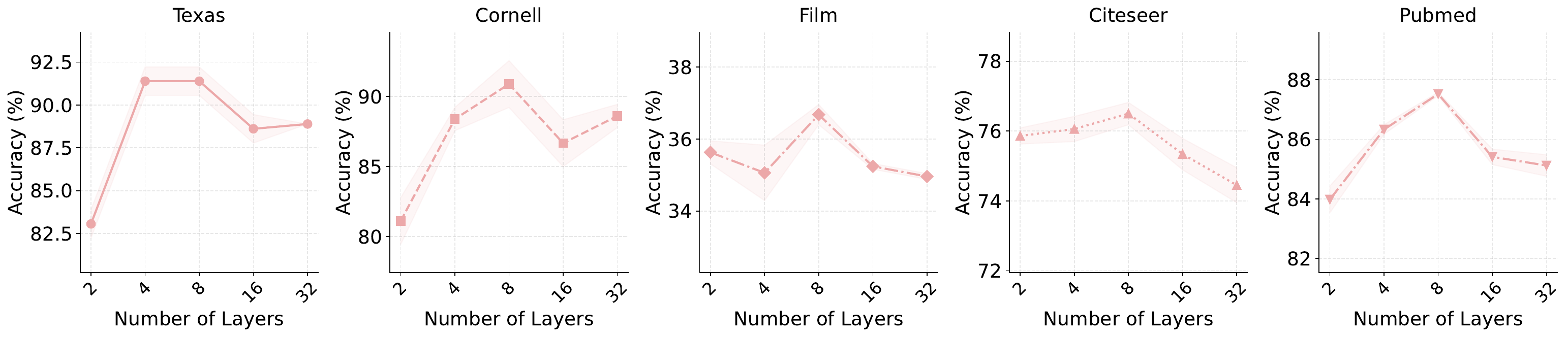}
\end{minipage}
\caption{Parameter sensitivity: Performance curves on five datasets with layers changing in $\{2, 4, 8, 16, 32\}$.}
\label{fig:layer}
\end{figure}

Figure \ref{fig:layer} explores the effect of network depth $L$.
For small graphs (Texas, Citeseer and Film), performance improves as the number of layers increases, since deeper networks are required to capture sufficient high-order information.
In contrast, for large graphs (Tolokers and Flickr), the best performance is achieved with only two layers, indicating that shallow message passing already provides sufficiently discriminative representations.
However, while APPNP can alleviate over-smoothing to some extent, it does not explicitly address this issue on these graphs; overcoming depth-related bottlenecks therefore remains an open direction for future research.

\section{Broader Impact Statement}\label{sec:limi}
This study aims to enhance message passing in graph neural networks through graph rewiring.
As a result, it contributes to better performance and broader applicability of GNNs across a wide range of tasks, including recommendation systems, molecular property prediction, traffic forecasting, and social network.
% , thereby improving the performance and practicality of GNN-based models  across various application scenarios.

% THR depends on sorting operations that impose considerable computational and memory burdens on large, densely connected graphs. Additionally, by assigning equal weight to distance and energy metrics, THR may inadvertently prune edges that carry essential semantic information and exhibit low interference when the distance is too large. 
% How to optimally balance these two metrics in accordance with the characteristics of different datasets remains an open challenge.
% Although our edge removal strategy is automated, the fixed number of edges added can lead to the introduction of spurious connections or the omission of beneficial ones. Hence, developing an adaptive edge addition approach is also crucial.
\bibliographystyle{ieeetr} 
\bibliography{sample-base}

\end{document}